\documentclass{article}

\usepackage{PRIMEarxiv}
\usepackage{microtype}
\usepackage{graphicx}
\usepackage{subfigure}
\usepackage{booktabs} % for professional tables
\usepackage{xcolor}
\usepackage{physics}
\usepackage{amsfonts}
\usepackage{amssymb}
\usepackage{bbm}
\usepackage{cite}
\usepackage[ruled]{algorithm2e}
\usepackage{bm}
\usepackage{amsmath,amssymb,amsfonts}
\usepackage{algorithmic}

\usepackage[T1]{fontenc}

\usepackage[ruled]{algorithm2e}
\usepackage{bm}

\usepackage{longtable}
\usepackage{cite}
\usepackage{amsmath,amssymb,amsfonts}
\usepackage{algorithmic}
\usepackage{graphicx}
\usepackage{textcomp}
\usepackage{xcolor}
\usepackage{bm}
\def\BibTeX{{\rm B\kern-.05em{\sc i\kern-.025em b}\kern-.08em
    T\kern-.1667em\lower.7ex\hbox{E}\kern-.125emX}}

\usepackage{amsthm}
\usepackage{multirow}
\newtheorem{prop}{Proposition}

\usepackage{caption}
\usepackage{url}            % simple URL typesetting
\usepackage{booktabs}       % professional-quality tables
\usepackage{amsfonts}       % blackboard math symbols
\usepackage{nicefrac}       % compact symbols for 1/2, etc.
\usepackage{microtype}      % microtypography
\usepackage{lipsum}
\usepackage{fancyhdr}       % header
\usepackage{graphicx}       % graphics
\graphicspath{{media/}}     % organize your images and other figures under media/ folder

\usepackage{orcidlink}

\usepackage{hyperref}

\hypersetup{
  colorlinks=false,
  linkbordercolor=white,
 urlbordercolor=white,
pdfborder={0 0 0}
}

\title{InQMAD: Incremental Quantum Measurement Anomaly Detection
\thanks{\textit{\underline{Citation}}: 
\textbf{Joseph et al., InQMAD: Incremental Quantum Measurement Anomaly Detection.}} 
}

\author{
  Joseph A. Gallego-Mejia \orcidlink{0000-0001-8971-4998}, Oscar A. Bustos-Brinez \orcidlink{0000-0003-0704-9117} , Fabio A. Gonz\'{a}lez\orcidlink{0000-0001-9009-7288} \\
  MindLab \\
  Universidad Nacional de Colombia \\
  Bogot\'{a}, Colombia\\
  \texttt{\{jagallegom,oabustosb,fagonzalezo\}@unal.edu.co} 
}

\begin{document}
\maketitle

\begin{abstract}
Streaming anomaly detection refers to the problem of detecting anomalous data samples in streams of data. This problem poses challenges that  classical and deep anomaly detection methods are not designed to cope with, such as conceptual drift and continuous learning. State-of-the-art flow anomaly detection methods rely on fixed memory using hash functions or nearest neighbors that may not be able to constrain high frequency values as in a moving average or remove seamless outliers and cannot be trained in an end-to-end deep learning architecture. We present a new incremental anomaly detection method that performs continuous density estimation based on random Fourier features and the mechanism of quantum measurements and density matrices that can be viewed as an exponential moving average density. It can process potentially endless data and its update complexity is constant $O(1)$. A systematic evaluation against 12 state-of-the-art streaming anomaly detection algorithms and using 12 streaming datasets is presented. 
\end{abstract}

\vspace{8pt}


\vspace{10pt}
\section{Introduction}
\label{sec:introduction}

Anomaly detection is a well-studied problem \cite{Chandola2009, Ruff2021, Pang2022}. The main idea is to detect data points or a group of data points that deviate from a `normality' in a specific context (note that normality is not related to the Gaussian normal distribution) . This problem arises in several domains, such as network security \cite{Mirsky2018}, telecommunications \cite{fernandes2019comprehensive}, retail industry \cite{nguyen2021forecasting}, network traffic \cite{wang2019detection}, financial transactions \cite{ahmed2016survey}, and wired and wireless sensors \cite{xie2011anomaly}. In recent years, particular interest has been given to methods that can deal with problems where data is continuously generated as a stream rather than as a batch of data points. This behavior is natural in credit card fraud detection \cite{Popat2018}, network system intrusion detection \cite{Mirsky2018}, camera surveillance \cite{sodemann2012review}, Internet of Things (IOT) device problems \cite{martiluis}, among others. Data in this streaming environment is rapidly generated, potentially infinite, has tremendous volume, and can exhibit concept drift. 

Classical anomaly detection algorithms in batch environments are based on density estimation, such as kernel density estimation \cite{latecki2007outlier}, classification, such as one-class support vector machines \cite{manevitz2001one}, and distance, such as the isolation forest \cite{liu2008isolation}. Another type of recent solutions are based on deep neural networks that have shown good properties in the anomaly detection task. They are mainly based on variational autoencoders \cite{an2015variational}, deep belief networks \cite{kwon2019survey}, one-class deep networks \cite{chalapathy2018anomaly} and adversarial autoencoders \cite{dimokranitou2017adversarial}. These methods assume that all training data points are available in the training phase. However, in a streaming context the data arrives continuously. Some of these methods have poor adaptability and extensibility, or inability to detect new anomalies continuously, where they have high model update cost and/or slow update speed.

To pave the way, several methods have been developed in the last decade. Methods such as iForestASD \cite{Ding2013}, RCF \cite{Guha2016}, xStream \cite{Manzoor2018} and Ex. IF \cite{Hariri2018} present a modification of the Isolation forest batch anomaly detection method. One of the problems of these methods is how to improve their inference complexity which is related to the depth of the trees which typically will be $log(n)$ where $n$ is the number of data points. To solve this problem, HS-Tree \cite{tan2011fast}, RS-Hash \cite{sathe2016subspace} and MStream \cite{Bhatia2021} use a hash structure to avoid traversing each tree. Other state-of-the-art methods use a modification of the well-studied local outlier factor (LOF) \cite{breunig2000lof}. Methods such as LODA \cite{tomaspvny}, DILOF \cite{Na2018} and MemStream \cite{Bhatia2021} use a modification of the $k$-nearest neighbor algorithm, the roots of LOF, to score the outlierness of data points. However, some of them are based on a memory that stores $m$-data points which, viewed as a moving average, may not be able to detect high frequency points or possible outliers.

In this paper we present the novel method incremental quantum measurement anomaly detection (InQMAD). This method uses adaptive Fourier features to map the original space to a Hilbert space. In this space, a density matrix is used to capture the density of the dataset. A new point passes through each stage and provides a score that is used as an anomaly score in the final stage. An important feature of the method is that it is able to build a density estimation model that is continuously updated with the incoming data and it has the ability to give more importance to recent samples similar to an exponential moving average. The method works in streaming, is an unsupervised algorithm and retraining can be performed continuously. The proposed method requires constant memory to process new data points, can process data in a single pass, and process potentially endless streaming data or even massive datasets. It can update the model in constant time, and its complexity is $O(1)$. 

In summary, the contributions of the present work are:

\begin{itemize}
    \item \textbf{A novel streaming anomaly detection method}: the new method works in a streaming, potentially infinite and with potentially concept drift environment.
    \item \textbf{A systematic evaluation of the proposed method}: the algorithm is evaluated in 12 streaming datasets and compared it against 12 state-of-the-art streaming anomaly detection methods.
    \item \textbf{An ablation study analyzing the new method}: a systematic evaluation of each component of the method is performed.
\end{itemize}

Reproducibility: the code used in this paper is released as open source and can be found in https://github.com/Joaggi/Incremental-Anomaly-Detection-using-Quantum-Measurements/ and zenodo https://doi.org/10.5281/zenodo.7183564

The outline of the paper is as follows: in Section 2, we describe anomaly detection in streaming and present the benchmark methods with which we will compare our algorithm. In Section 3, we present the new method, explaining all the stages of the algorithm. In Section 4, we systematically evaluate the proposed method against state-of-the-art streaming anomaly detection algorithms. In Section 5, we state conclusions and outline future lines of research.

\vspace{10pt}
\section{Background and Related Work}

\vspace{5pt}
\subsection{Streaming Anomaly Detection}
An anomaly can be broadly defined as an observation or data that deviates significantly from some kind of normality. Several types of anomalies can occur in real datasets, such as point anomalies, group anomalies, contextual anomalies, low-level texture anomalies, and high-level semantic anomalies. Classical methods have been used to solve this problem, but they suffer with high-dimensional datasets \cite{kwon2019survey}. Most of the recent deep learning methods capture a lot of attention for their good properties, such as automatic feature extraction. However, their training time and streamwise inference is prohibitively long \cite{Bhatia2022}.

Stream anomaly detection can be viewed as a generalization of the typical anomaly detection problem where data grows infinitely. Therefore, it is impractical, impossible or unnecessary to store every data point that arrives as a stream. In this context, the method has to distinguish between normal and anomalous data points, where concept drift can occur and the number of anomalies is scarce compared to normal data \cite{Ding2013}. Several types of concept drift can arise in streaming datasets, such as sudden, gradual, incremental or recurrent drift. Concept drift occurs in data streams, where usually old data is less important than new data. This trend in data has an evolutionary pattern, where recent behavior should be of greater importance than older patterns \cite{tomaspvny}. In order to solve this problem, the method must use a constant memory and a nearly constant inference processing time. Therefore, it will process the data in a single pass. In the following subsection \ref{subsec:baseline_methods}, 12 methods for streaming anomaly detection are presented.

\subsection{Streaming Anomaly Detection Problem}

In streaming anomaly detection, the data points $\mathcal{X}=\{\bm{x}_1, \cdots, \bm{x}_t\}$ arrive as a $d$-feature-dimension sequence. This sequence can be the sequence of transactions for a given credit card or the temperature recorded by an IOT sensor. The challenge in this configuration in that the concept of "normality" evolves over time, meaning that a point that was considered normal behavior can drift and become anomalous behavior. To solve this problem, there is a need to develop algorithms that can learn on-line with high speed and low memory consumption.
\vspace{5pt}
\subsection{Streaming Anomaly Detection Baseline Methods}\label{subsec:baseline_methods}

\begin{itemize}
\item STORM \cite{Angiulli2007}: A Stream Manager and a Query Manager have been proposed. The former is an indexed stream buffer whose job is to store the number of successive neighbors and the identifiers of the most recent preceding neighbors. The latter is a procedure that efficiently answers queries about whether a certain data point is an inlier or an outlier.

\item HS-Tree \cite{tan2011fast}: This method constructs complete binary trees where each tree has at most $2^{h+1} - 1$ node. Each subtree is constructed by randomly selecting a feature and breaking it in half. A random perturbation of each subtree is performed to create diverse subtrees. Each point has to traverse each binary tree to capture the mass profile. Finally, a scoring function is computed using the mass function of a query data point passing through each data node.

\item iForestASD \cite{Ding2013}: This algorithm has its roots in the Isolation Forest method. The method uses a window of the streaming data and is sent to the Isolation Forest. An abnormality score is calculated using the average of the depth of the point on each tree in the forest. If the point is normal, it is joined to the Isolation Forest.

\item RS-Hash \cite{sathe2016subspace}: This method uses the Isolation Forest method at its roots. A tree is constructed as a sequence of randomly selected features. Anomalous points are those that are easily separated from the normal data points. With the scoring function given by the Isolation Forest, a score of anomalous sliding windows is computed to detect concept drift.

\item RCF \cite{Guha2016}: The algorithm constructs a robust randomized cutting tree (rrct) using a random selection of the dimension weighted by its range. A uniform distribution is used to cut the selected dimension. This process is repeated $n$ times, with $n$ being the maximum depth. A forest is constructed using several rrct. A new point is classified as an anomaly using the comparison of its insertion and deletion complexity. 

\item LODA \cite{tomaspvny}: The algorithm uses a set of histograms for each dimension. Each histogram is mapped to a projection space using $\bm{w}$ parameters that capture the importance of the feature. The log likelihood in the projection space is then calculated. An anomalous data point is expected to have a lower value of the log likelihood.

\item Kitsune \cite{Mirsky2018}: Kitsune is an algorithm that uses an ensemble of autoencoders to provide an anomaly score. Features are sent to $l$ autoencoders of 3 layers each. The reconstruction error calculated as root mean square error (RMSE) is sent to a final 3-layer autoencoder. Finally, the RMSE of the reconstruction is calculated and used as the anomaly score.

\item DILOF \cite{Na2018}: The method uses the $k$ nearest neighbor information as in the Local Outlier Factor (LOF) but improves it in the case of streaming data. The algorithm has two phases: a detection phase and a summary phase. In the first phase it decides whether a point is anomalous or not. In the second, it uses an approximation algorithm with a sampling strategy to update the memory of the $k$ nearest neighbors.

\item xStream \cite{Manzoor2018}: The method uses a hash structure to reduce the dimensionality of the data points, which allows the evolution of features in the stream. After reduction, the method partitions the space into so-called half-space chains. These partitions capture the density estimate at a different granularity. The outlier score is calculated based on the density estimate score. For streaming, a previous window is used to score the point outlier. 

\item MStream \cite{Bhatia2021}: This algorithm uses two locality-sensitive hash functions: feature hashing and record hashing. The former computes a hash for each feature in the data point. The latter computes a hash for all features simultaneously.  The anomaly score is calculated using a chi-square density function. The algorithm is combined with an autoencoder to reduce the dimensionality of the data.

\item Ex. IF \cite{Hariri2018}: This algorithm uses a random slope to cut the hyperplane and a random intercept. This differs from the isolation forest because the latter chooses random features and generates a random cut on that feature. The method shows better results when compared to the isolation forest for anomaly detection.

\item MemStream \cite{Bhatia2022}: The method proposes a nearest neighbor memory based algorithm. Each data point passes through a shallow autoencoder to reduce the dimensionality of the original space. A memory is built using $n$-normal data points as initialization. Then, when a new point arrives it is forwarded to the autoencoder and compared to the memory. An anomalous point will have a high mean square error compared to its neighbors. The memory is updated only with normal data points.

\end{itemize}

\vspace{10pt}
\section{Incremental Quantum Measurement Anomaly Detection (InQMAD)}\label{section:anomaly_detection_method}

\begin{figure*}[t]
\begin{centering}
\includegraphics[scale=0.56]{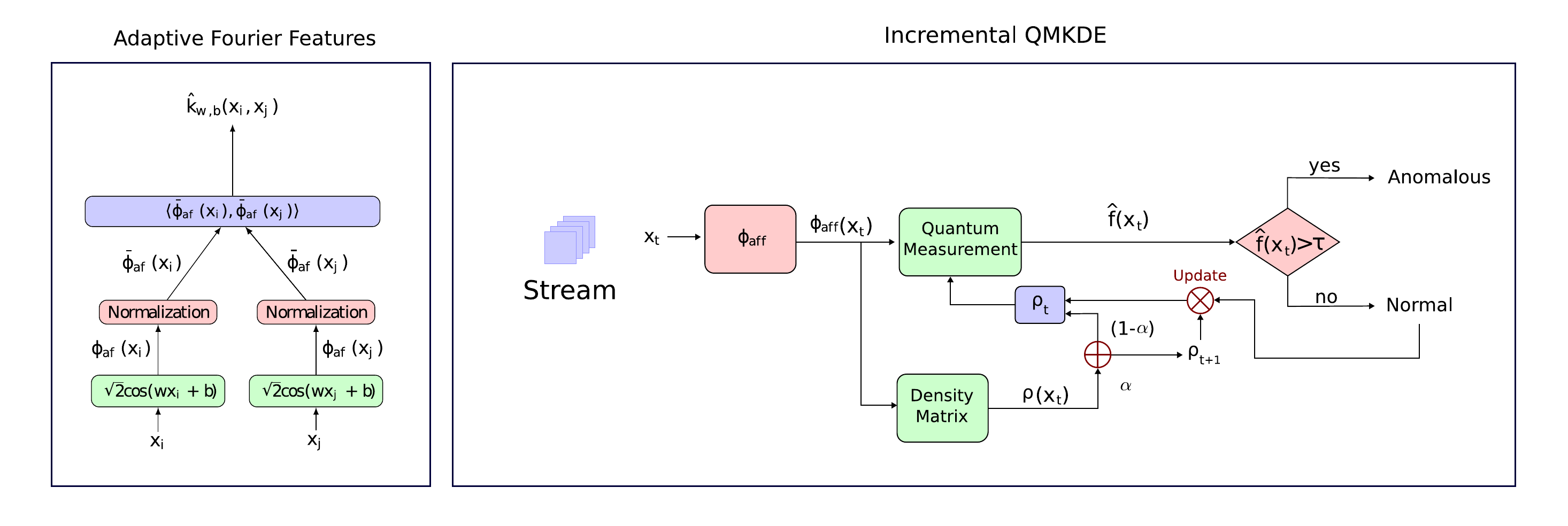}
\par\end{centering}
\caption{Incremental Quantum Measurement Anomaly Detection (InQMAD) method. The method consists of five steps: (1) an adaptive Fourier features (left subfigure), (2) a density matrix initialization, (3) a quantum measurement of the new streaming data points, (4) a decision anomaly detection threshold for the new data, and (5) an update of the density matrix $\rho$ in case of new normal data points. \label{fig:model}}
\end{figure*}

In this section, we present the new method new Incremental Quantum Measurement Anomaly Detection (InQMAD). Figure \ref{fig:model} shows each of the steps of the Algorithms \ref{alg:algorithm_initalization} and \ref{alg:algorithm_inference}. The method consists of five steps: (1) an adaptive Fourier features, (2) a density matrix initialization, (3) a quantum measurement of the new streaming data points, (4) a decision anomaly detection threshold for the new data, and (5) an update of the density matrix $\rho$ in case of new normal data points. Each stage is explained in detail in this section.

\subsection{Adaptive Fourier Features}

All streaming data points are mapped into a Hilbert space using adaptive Fourier features, first proposed as random Fourier features by \cite{rahimi2007rff}. They showed that a Gaussian kernel can be approximated as: 

\begin{equation} \label{eq:randomized_map}
    k(\bm{x}, \bm{y}) \simeq \mathbb{E}_{\bm{w}}(\langle \hat{\phi}_{\text{rff}, \bm{w}}(\bm{x}), \hat{\phi}_{\text{rff}, \bm{w}}(\bm{y})\rangle )
\end{equation}

where  $\hat{\phi}_{\text{rff}, \bm{w}} = \sqrt{2} \cos (\bm{w}^T\bm{x} + b)$,  $\bm{w} \sim N(\bm{0},\mathbb{I}^d)$ and $\bm{b} \sim \text{Uniform}(0, 2\pi)$. The expectation is possible to Bochner's Theorem and the fact that using Equation \ref{eq:randomized_map} that defines a randomized map converge in probability to the Gaussian kernel. However, the randomized map can be further refined using an optimization step, in particular using gradient descent, as shown in \cite{li2019learning, gallego2022neural}. Given two pairs of random data points ($\bm{x}_i$, $\bm{x}_j$), the parameters $\bm{w}, b$ can be optimized using the following equation:

$$\bm{w}^*, b^* = \arg \min_{\bm{w},b} \frac{1}{m}\sum_{\bm{x}_i,\bm{x}_j\in s} (k_\sigma(\bm{x}_i,\bm{x}_j) - \hat{k}_{\bm{w},b}(\bm{x}_i,\bm{x}_j))^2$$
where $\hat{k}_{\bm{w},b}(\bm{x}_i,\bm{x}_j))^2$ is the Gaussian kernel of Fourier feature approximation and $k_\sigma(\bm{x}_i,\bm{x}_j)$ is the Gaussian kernel with a bandwidth parameter $\sigma$. Adaptive Fourier features allow us to compute the feature space avoiding the explicit kernel computation and will be used as input to the density matrix in the next step. We performed an empirical evaluation of this improvement algorithm and propose an intermediate enhancement step. Assume that all points are normalized between $[0,1]^d$, where $d$ is the number of features, generate random samples from $\text{Uniform}(-0.5,1.5)^d$. The increase in the range of the sampling space is to account for future points outside the range. The next question that arises is how many points we should generate. We use an empirical approach depending on how large the initial training dataset is. If the training set is small ($<$ 1000), we will sample until it consists of 10,000 data points. Otherwise, we will sample twice the size of the initial training dataset. The intuition here is that if we have few data points, the algorithm will be prone to overfitting in a local space near initial training dataset. The algorithm \ref{alg:algorithm_initalization} shows the step in \ref{alg:aff}. 

\subsection{Density Matrix Initialization}\label{subsec:dmi}

The second step of the algorithm consists of calculating the density matrix using the mapping obtained by passing each $\bm{x}_t$ to the adaptive Fourier feature step explained above. The density matrix is a formalism of quantum mechanics that was used as a base tool by \cite{gonzalez2021learning,gallego2022neural} to create a density estimation method. The authors derive a new algorithm that uses random Fourier features to store the density matrix $\rho$. The following equation is a slight modification of the density matrix in terms of stream data: 

\begin{equation} \label{eq:density_matrix}
  \rho_t = \frac{1}{n}\sum_{i=1}^N q_i \cdot \phi_\text{aff}(\bm{x_i})\phi^t_\text{aff}(\bm{x_i})
\end{equation}

where $\sum_{i=1}^{t} q_i = 1$. To initially compute the density matrix, an initial portion of the stream $\{\bm{x}_1, \cdots, \bm{x}_n\}$ is selected and sent to the equation \ref{eq:density_matrix}
to calculate the density matrix $\rho$. The initial training dataset size can degrade the final performance of the algorithm; therefore, it is necessary to find it using a cross-validation approach. It should be noted that the density matrix is computed using adaptive Fourier features avoiding explicit kernel computation as in kernel density estimation; however, other feature mappings can be used as shown in \cite{gonzalez2021classification}. The density matrix is initialized in Algorithm \ref{alg:algorithm_initalization} in Step \ref{alg:rho}.

\subsection{Quantum Measurement}

The third step of the algorithm is the quantum measurement. A quantum measurement can be obtained by using the density matrix $\rho$ and mapping a data point $\bm{x}_{t+1}$ at time $t+1$ as: 

\begin{align} \label{eq:kde_density_matrix}
\begin{aligned}
\hat{f}(\bm{x}_{t+1}) & = \frac{1}{M_\sigma} {\phi}_\text{aff}(\bm{x}_{t+1})^T \rho_t \: {\phi}_\text{aff}(\bm{x})
\end{aligned}
\end{align}
where $\rho$ is the density matrix defined in the equation \ref{eq:density_matrix} and $M_\sigma$ is a normalization constant. This step gives us an estimate of the density of the given data point that will be used in the next step. The initialized Algorithm \ref{alg:algorithm_initalization} uses the quantum measurement in Step \ref{alg:quantum_measurement} and the inference Algorithm \ref{alg:algorithm_inference} uses it in Step \ref{alg:quantum_measurement_inference}.

\vspace{5pt}
\subsection{Anomaly Detection Classification}\label{subsec:threshold_calculation}

The fourth step of the algorithm is the anomaly detection classification stage. The threshold is defined as $\tau$, where a new point $\bm{x}_i$ is defined as anomalous according to the following equation:

$$\hat{y}(\bm{x}_{i}) = \left\{\begin{matrix}
 \text{ `normal'} & \textit{if } \hat{f}(\bm{x}_{i}) \geq \tau\\ 
 \text{ `anomaly'} & \textit{otherwise}
\end{matrix}\right.$$

After initialization of the density matrix using the initial training dataset, the threshold is found using two different approaches. The first is that the threshold can be found using prior knowledge of the proportion of anomalies ($\beta$). The second approach is to use an optimization metric computed over $\{\hat{f}(\bm{x_1}), \cdots, \hat{f}(\bm{x_n})\}$, for instance using the best threshold with respect to AUC-ROC in initial memory. The $\tau$ is obtained in Step \ref{alg:initialization_tau} of the Algorithm \ref{alg:algorithm_initalization} and the anomaly detection classification occurs in Step \ref{alg:classification_inference} of the Algorithm \ref{alg:algorithm_inference}.

\subsection{Density Matrix Update}

The fifth and final step of the algorithm is to update the density matrix. If a new point $\bm{x_t}$ at time $t$ of the flow is classified as a normal data point, the density matrix $\rho_{t+1}$ will be calculated as follows:

\begin{align} \label{eq:density_matrix_update}
  \rho_{t+1} &=  (1-\alpha) \cdot \rho_{t} + \alpha \cdot \phi_\text{aff}({\bm{x}_{t+1}})\phi^T_\text{aff}({\bm{x}_{t+1}})
\end{align}

\begin{prop} \label{eq:prop}
The resulting matrix $\rho_{t+1}$ in Equation \ref{eq:density_matrix_update} is a valid density matrix of the form $\rho_{t+1} = \sum_{i=1}^{t+1} q_i \phi({\bm{x}_{i}})\phi^T({\bm{x}_{i}})$ with $q_1 =  (1-\alpha)^{t+1}$, $q_i = (1-\alpha)^{t-i+1} \cdot \alpha \text{ } \forall i \in \{1,\cdots, t\}$ and $\alpha \in [0,1]$.
\end{prop}

 \begin{figure}[t]
\begin{centering}
\includegraphics[scale=1]{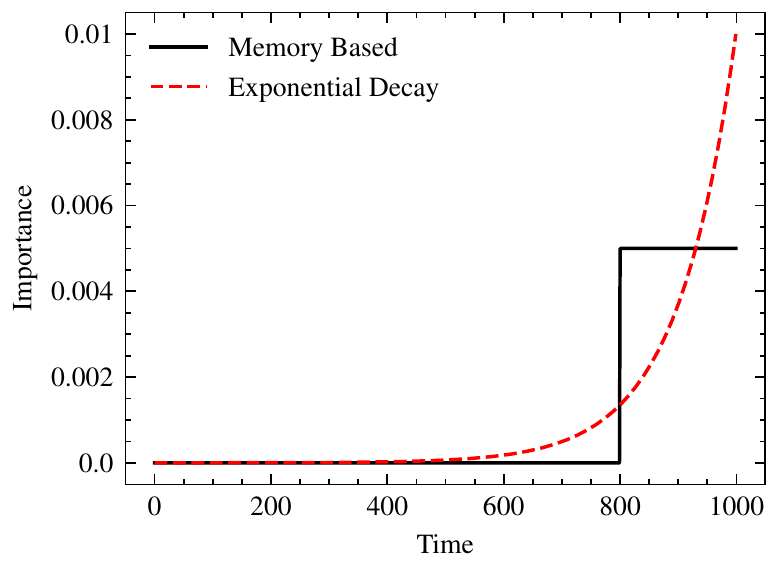}
\par\end{centering}
\caption{Importance comparison between memory based against exponential decay\label{fig:importance}.}
\end{figure}

Proposition \ref{eq:prop} shows that an exponential decay defines a valid density matrix. This matrix consists of all data points that have reached the method up to time $t$, where older examples are less important and have a lower weight compared to more recent ones similar to an exponential moving average.

Figure \ref{fig:importance} shows the difference between a constant memory and an exponential decay method. The figure shows a memory-based algorithm using a window width of 200 and compares it against an exponential decay memory using a $\alpha = \frac{2}{\text{memory window width}}$. The black memory-based line shows a constant memory whose points are of equal importance to the method. The red exponential decay line shows an exponentially decaying method whose points become less and less important as they age. The density matrix update only occurs if the new point is classified as `normal' and is performed in Step \ref{alg:density_matrix_update} of Algorithm \ref{alg:algorithm_inference}.

\subsection{Complexity Analysis}\label{subsec:threshold_calculation}

%Theorem : The density update defined by Equation \ref{eq:density_matrix_update} is a valid density matrix
%
%Define an operator $\hat{O}$ and compute their expectation value as follows:
%
%\begin{align} \label{eq:density_matrix_update}
%  \langle \hat{O} \rangle &= (1-\alpha) \cdot \frac{1}{t}\sum_{i=1}^t \langle \psi_i | \hat{O} | \psi_i \rangle + \alpha \cdot \langle \psi_{t+1} | \hat{O} | \psi_{t+1} \rangle\\
%  &= (1-\alpha) \cdot \frac{1}{t}\sum_{i=1}^t Tr(\hat{O} |\psi_i\rangle \langle \psi_i |) + \alpha \cdot Tr(\hat{O} |\psi_{t+1}\rangle \langle \psi_{t+1} |) \label{eq:density_matrix_trace}\\
%    &= Tr\left(\hat{O} \left((1-\alpha) \cdot \frac{1}{t}\sum_{i=1}^t  |\psi_i\rangle \langle \psi_i | + \alpha \cdot |\psi_{t+1}\rangle \langle \psi_{t+1} |\right)\right)
%\end{align}
%
%\begin{align} \label{eq:density_matrix_update}
%    &= Tr\left(\hat{O} \left((1-\alpha) \cdot \frac{1}{t} \left[ |\psi_1\rangle \langle \psi_1 | + \cdots + |\psi_t\rangle \langle \psi_t | \right + \alpha \cdot |\psi_{t+1}\rangle \langle \psi_{t+1} |\right)\right)
%\end{align}
%
%The equation \ref{eq:density_matrix_trace} is thanks to cyclic invariance and linearity properties of the trace. The In order to the last expression be a valid density matrix, the sum of each coefficient inside the trace must be one. Then:
%
%\begin{align} \label{eq:density_matrix_update}
%    1 &= (1-\alpha)\cdot \frac{1}{t} \cdot t + \alpha \\
%     &= (1-\alpha)\cdot \frac{1}{t} \cdot t + \alpha \cdot \frac{t}{t} \\
%    &= \frac{t}{t} -\alpha \cdot \frac{ t}{t} + \alpha \cdot \frac{t}{t} \\
%    &= 1
%    \end{align}

In this subsection, $d$ will be the number of features and $D$ will be the number of adaptive Fourier features. For time complexity, in the anomaly detection phase, a dot product is computed between each data point and the adaptive Fourier feature whose time complexity is proportional to $O(dD)$. In addition, a dot product is computed from the above result and the density matrix whose time complexity is proportional to $O(D^2)$. In terms of memory, InQMAD needs to maintain a density matrix whose size is proportional to $O(D^2)$ and stores the weights of the adaptive Fourier features whose size is proportional to $O(dD)$.

\begin{algorithm}[tb]
\caption{InQMAD initialization process}
\label{alg:algorithm_initalization}
\textbf{Input}: Training dataset $D=\{\bm{x}_t\}_{t=1,\cdots,n},  \bm{x}_t \in \mathbb{R}^d$ \\
$\alpha$: forgetting trade-off,\\
$\sigma$: bandwith parameter\\
$\beta$: proportion of anomalies\\
\textbf{Output}: $\bm{w}_{\text{AFF}}, b_{\text{AFF}}, \rho, \tau$ \\
\begin{algorithmic}[1] %[1] enables line numbers
\STATE $\bm{w}_{\text{AFF}}^*, b_{\text{AFF}}^* = \arg \min_{\bm{w},b} \frac{1}{m}\sum_{\bm{x}_i,\bm{x}_j\in D}(  k_{\sigma}(\bm{x}_i,\bm{x}_j)-\hat{k}_{\bm{w},b}(\bm{x}_i,\bm{x}_j))^2$ \label{alg:aff}
\FOR{$\bm{x}_t \in D $}
\STATE $\rho_t = (1-\alpha) \cdot \rho_{t-1} +  \alpha \cdot \phi_\text{aff}({\bm{x}_t})\phi^T_\text{aff}({\bm{x}_t})$\\ \label{alg:rho}
\STATE $\hat{f}(\bm{x}_{t}) = \frac{1}{M_\sigma} {\phi}_\text{aff}(\bm{x}_{t})^T \rho_t \: {\phi}_\text{aff}(\bm{x}_{t})$ \label{alg:quantum_measurement}
%\IF {conditional}
%\STATE Perform task A.
%\ELSE
%\STATE Perform task B.
%\ENDIF
\ENDFOR 
\STATE $\tau = q_{\beta} (\hat{f}(\bm{x}_1), \cdots, \hat{f}(\bm{x}_n))$ \label{alg:initialization_tau}

\STATE \textbf{return} $\bm{w}_{\text{AFF}}, b_{\text{AFF}}, \rho_n, \tau$
\end{algorithmic}
\end{algorithm}

\begin{algorithm}[tb]
\caption{InQMAD inference and density matrix update}
\label{alg:algorithm_inference}
\textbf{Input}: $\bm{x}_{t+1}$ \\
$\rho_{t}$: density matrix\\
$\alpha$: forgetting trade-off,\\
$\bm{w}_{\text{aff}},b_{\text{aff}}$: adaptive Fourier features parameters\\
$\tau$: threshold anomaly detector\\

\textbf{Output}: $\rho_{t+1}$-quantum measurement KDE parameter, \\
$\hat{y}_{t+1}$ classification of the given data point  \\
\begin{algorithmic}[1] %[1] enables line numbers
\STATE $\hat{f}(\bm{x}_{t+1}) = \frac{1}{M_\sigma} {\phi}_\text{aff}(\bm{x}_{t+1})^T \rho_t \: {\phi}_\text{aff}(\bm{x}_{t+1})$ \label{alg:quantum_measurement_inference}
\IF { $\hat{f}(\bm{x}_{t+1}) \geq \tau$} \label{alg:classification_inference}
\STATE $\rho_{t+1} = (1-\alpha) \cdot \rho_{t-1} +  \alpha \cdot \phi_\text{aff}({\bm{x}_{t+1}})\phi^T_\text{aff}({\bm{x}_{t+1}})$ \label{alg:density_matrix_update}
\STATE $\hat{y}_{t+1} = `normal'$
\ELSE
\STATE $\rho_{t+1} = \rho_{t}$
\STATE $\hat{y}_{t+1} = `anomaly'$
\ENDIF
\STATE \textbf{return} $\rho_{t+1}, \hat{y}_{t+1}$
\end{algorithmic}
\end{algorithm}

\vspace{10pt}
\section{Experimental Evaluation}
\label{sec:experimental_evaluation}

We designed and tested an experimental setup in order to answer the following questions:

\begin{itemize}
    \item \textbf{Q1. Streaming Method Comparison.} Does our method performs accurately in anomaly detection over streams of data, when compared it against to state-of-the-art baseline methods?
    \item \textbf{Q2. Adaptability.} How well our method does adapt to ``concept drift", that is, sudden changes in the data stream inner structure?
    \item \textbf{Q3. Ablation Study.} What are the effects of removing some stages of the algorithm (in particular, the Adaptive Fourier features embedding) on the overall performance of our method?
\end{itemize}

\subsection{Comparison to Streaming Methods}

\subsubsection*{Experimental Setup}

The experimental setup presented in this paper took inspiration from the setup proposed in \cite{Bhatia2022}. There, it was performed a comparison of a dozen of algorithms, based on area under the ROC curve (commomly known as AUC or AUROC), by applying them over a series of streaming and anomaly detection datasets. Our method was implemented using the JAX framework, a Python library designed for high-performance machine learning. All the experiments were carried out on a machine with a 2.1GHz Intel Xeon 64-Core processor with 128GB RAM and two NVIDIA RTX A5000 graphic processing units, that run Ubuntu 20.04.2 operating system.

To handle the inherent randomness that the proposed method can present in some of its stages (particularly in dataset splitting and neural network training), we selected a unique, invariant value for all the random seeds that affect the behavior of our method.

\vspace{5pt}
\subsubsection*{Datasets}

In this framework, we selected twelve different datasets, that can be divided into two groups: seven datasets (Cardio, Ionosphere, Mammography, Pima, Satellite, Satimage and Synthetic) with relatively low dimensions and low number of records, mainly used to perform proof-of-concept in anomaly detection, and five datasets (KDD99, NSL, DoS, UNSW and Cover) with hundreds of thousands of registers and a high number of dimensions, that require an intensive use of resources to be processed. A summary of the main features of all the datasets can be seen in Table \ref{dataset_features}, and a brief description of each dataset is presented in the Supplemental Material.

\begin{table}[t]
\centering

\caption{Main features of the datasets}
\label{dataset_features}

\begin{tabular}[c]{l|c|c}
\hline
Dataset & Records & Dimensions \\ 
\hline

Cardio & 1831 & 21 \\
Ionosphere & 351 & 33 \\
Mammography & 11183 & 6 \\
Pima & 768 & 8 \\
Satellite & 6435 & 36 \\
Satimage & 5803 & 36 \\
Synthetic & 10000 & 1 \\
\hline
KDD99 & 494021 & 121 \\
NSL & 125973 & 126  \\
Cover & 286048 & 10 \\
DoS & 1048575 & 95\\
UNSW & 2540044 & 122  \\

\hline
\end{tabular}

\end{table}

\vspace{5pt}
\subsubsection*{Parameter Search}

The behavior of InQMAD depends on a series of parameters that regulate the Adaptive Fourier features embedding, the density estimation stage and the size of the initialization dataset of the method. Particularly, we tried to find which of these parameters had a larger impact over the performance of the algorithm, and we selected four to carry out a parameter grid search over them in order to find the combination of parameters that showed the best performance. The selected parameters are the following:

\begin{itemize}
    \item $n$: this corresponds to the size of the initial training dataset $D$, that is used in the initialization stage of the method. The $n$ samples allow the construction of the first density matrix $\rho_n$, and thus they indirectly influence the subsequent behavior of the method. We searched this value in the set $\left\{ 64,128,256,512,1000, 2000,2048,5000 \right\}$, taking into account that it cannot be bigger than the total of samples in the dataset.
    \item $lr_{base}$: since the construction of the adaptive Fourier features parameters $\bm{w}_{\text{aff}}$ and $b_{\text{aff}}$ involves the training of a neural network, we searched the best possible values for the learning rate of this process. We designed the learning rate to follow a polynomial decay, so we needed a start point and an end point for it. With the end point ($lr_{end}$) fixed on the value $10^{-7}$, we searched the start point ($lr_{base}$) over the values \{$10^{-2}$, $10^{-3}$, $10^{-4}$\}.
    \item $\sigma$ (of kernel): we used a Gaussian kernel at the core of the KDE stage of the method. The shape of this kernel depends on the variance value, commonly notated as $\sigma^2$, and it has a notable influence over the quality of the density estimation. Since the variance is related to the structure of the data, we chose a different set of values for each dataset. 
    \item $\alpha$: this parameter controls the tradeoff between the samples that are stored into the density matrix of the method and the incoming samples that can substitute them. This parameter can only take values between zero and one, and bigger values correspond to bigger substitution rates on the memory. We searched this parameter over the range $[0.001, 1)$. 
\end{itemize}

The winning combinations of parameters for each dataset can be found in the Supplemental Material. 

\vspace{5pt}
\subsubsection*{Evaluation Metrics}

Following the framework established in \cite{Bhatia2022}, we selected the area under the ROC curve (also known as AUC or AUC-ROC) as the main metric to establish the performance of our algorithm and compare it with the baseline methods. This metric was chosen because it represents the overall capability of the model to distinguish between the classes, regardless of the thresholds used in specific situations.  

%Besides, we also implemented into our method the calculation of other metrics such as area under the Precision-Recall curve (AUC-PR), accuracy and weighted F1-Score.

\vspace{5pt}
\subsubsection*{Results and discussion}

The AUCROC score obtained for every pair of dataset and algorithm is presented in Table \ref{results}. The metrics corresponding to our method can be found in the column labeled InQMAD. The absent values correspond to cases where the algorithms were not able to handle that particular dataset. For each row, the best value is marked in bold, the second is underlined and the third is marked in italics.

When looking at the average performance, InQMAD is the best method, being slightly better than MemStream, and notably better than all other methods, thus showing a noticeable improvement over the state-of-the-art algorithms in the area. When considering the size of the datasets, there is a clear advantage of InQMAD over the smaller datasets (the ones in the top rows of Table I), due to the fact that it shows the best performance for all of these datasets. For bigger datasets (the ones in the bottom rows of Table I), it is still competitive but some algorithms do better, mainly in the biggest datasets like DoS or UNSW. 

On the other hand, the number of dimensions does not seem to be as strongly related to the performance of our method as the size, considering that it had a mixed behavior for datasets with low dimensionality (high for Pima or Satellite, lower for Cover) and for datasets with high dimensions (high for KDD or NSL, lower for UNSW). In general, memory-based methods, such as MemStream, underperform InQMAD. Our intuition for this behavior is that memory-based algorithms are similar to moving averages that may be prone to a high frequency point or outlier. However, InQMAD can be considered an exponential moving average that dampens the weight of high frequency points.  

\begin{table*}[t]
\centering
\scriptsize
\caption{Area under the ROC curve (AUCROC) for all algorithms over all datasets. The first, second and third positions are respectively highlighted in bold, underlined and italics.}
\label{results}
\begin{tabular}{l|cccccccccccc|c}
\hline
Dataset & \scriptsize{STORM} & \scriptsize{HS-Tree} & \scriptsize{iForestASD} & \scriptsize{RS-Hash} & \scriptsize{RCF} & \scriptsize{LODA} & \scriptsize{Kitsune} & \scriptsize{DILOF} & \scriptsize{xStream} & \scriptsize{Mstream} & \scriptsize{Ex.IF} & \scriptsize{MemStream} & InQMAD \\
\hline
Cardio & 0,507 & 0,673 & 0,515 & 0,532 & 0,617 & 0,501 & \textit{0,966} & 0,570 & 0,918 & \underline{0,986} & 0,921 & 0,884 & \textbf{0,989} \\
Cover & 0,778 & 0,731 & 0,603 & 0,640 & 0,586 & 0,500 & 0,888 & 0,688 & 0,894 & 0,874 & \textit{0,902} & \textbf{0,952} & \underline{0,923} \\
DoS & 0,511 & 0,707 & 0,529 & 0,527 & 0,514 & 0,500 & \textit{0,907} & 0,613 & 0,800 & \underline{0,930} & 0,734 & \textbf{0,938} & 0,788 \\
Ionosphere & 0,637 & 0,764 & 0,694 & 0,772 & 0,675 & 0,503 & 0,514 & \textbf{0,928} & 0,847 & 0,670 & \textit{0,872} & 0,821 & \textbf{0,928} \\
KDD & 0,914 & 0,912 & 0,575 & 0,859 & 0,791 & 0,500 & 0,525 & 0,535 & \textit{0,957} & 0,844 & 0,874 & \underline{0,980} & \textbf{0,995} \\
Mammo & 0,650 & 0,832 & 0,574 & 0,773 & 0,755 & 0,500 & 0,592 & 0,733 & 0,856 & 0,567 & \textit{0,867} & \underline{0,894} & \textbf{0,914} \\
NSL & 0,504 & \textit{0,845} & 0,500 & 0,701 & 0,745 & 0,500 & 0,659 & 0,821 & 0,552 & 0,544 & 0,767 & \underline{0,978} & \textbf{0,982} \\
Pima & 0,528 & 0,667 & 0,525 & 0,562 & 0,571 & 0,502 & 0,511 & 0,543 & 0,663 & 0,529 & \textit{0,672} & \underline{0,742} & \textbf{0,750} \\
Satellite & 0,662 & 0,519 & 0,504 & 0,675 & 0,552 & 0,500 & 0,665 & 0,561 & 0,677 & 0,563 & \textit{0,716} & \underline{0,727} & \textbf{0,764} \\
Satimage & 0,514 & 0,929 & 0,554 & 0,685 & 0,738 & 0,500 & 0,973 & 0,563 & \textbf{0,996} & 0,958 & \textit{0,995} & 0,991 & \textbf{0,996} \\
Syn & 0,910 & 0,800 & 0,501 & \textit{0,921} & 0,774 & 0,506 & - & 0,703 & 0,539 & 0,505 & - & \underline{0,955} & \textbf{0,982} \\
UNSW & 0,810 & 0,769 & 0,557 & 0,778 & 0,512 & - & 0,794 & 0,737 & 0,804 & \textit{0,860} & 0,541 & \textbf{0,972} & \underline{0,873} \\
\hline
\end{tabular}
    
\end{table*}

\subsection{Ablation Study}

Being one of the most important stages of our algorithm, we wanted to determine the degree of influence of the Adaptive Fourier Features embedding over the algorithm performance. This comparison was made by building alternate versions of our method: InQM-NoAdp, that instead of using adaptive Fourier features as mapping, uses the random Fourier features approach stated in previous works like \cite{gonzalez2021learning}, and InQMAD-200, where the number refers to the size of the encoding given by adaptive Fourier features. Since the size of the encoding was chosen to be 2000 in all the experiments of InQMAD, we expect to determine if the size of the adaptive features encoding enhances or diminishes the overall performance of the method. 

Using these different versions of our method over the datasets in the experimental setup, we calculated the metrics that we present in Table \ref{ablationstudy}. In order to make a fair comparison, in every dataset we used the same parameters for all the versions. The best result of each row is marked in bold.

From these results, there is a clear advantage in using the adaptive Fourier Features in the majority of datasets, regardless of their size or dimensionality. Only in one of the datasets (DoS) there is a small decrease in performance when using many Adaptive features, in favor of a lower embedding size. Although, the difference in this case is way smaller than the differences in favor of the use of Adaptive features on other datasets, particularly Ionosphere, Syn and Satellite.

\begin{table}[ht!]
\caption{Results for Ablation Study on InQMAD, including Adaptive and NoAdaptive versions}
\label{ablationstudy}

\centering
\begin{tabular}{l|ccc}

\hline
\multirow{2}{*}{DATASET} & \multicolumn{3}{c}{AUCROC} \\ 
 & InQM-NoAdp & InQMAD-200 & InQMAD \\ \hline
Cardio & 0,976 & 0,974 & \textbf{0,989} \\
Cover & 0,847 & 0,899 & \textbf{0,923} \\
DoS & 0,767 & \textbf{0,808} & 0,788 \\
Ionosphere & 0,825 & 0,852 & \textbf{0,928} \\
KDD & 0,937 & 0,943 & \textbf{0,995} \\
Mammo & 0,907 & 0,912 & \textbf{0,914} \\
NSL & 0,947 & 0,96 & \textbf{0,982} \\
Pima & 0,728 & 0,737 & \textbf{0,75} \\
Satellite & 0,67 & 0,694 & \textbf{0,764} \\
Satimage & 0,965 & 0,977 & \textbf{0,996} \\
Syn & 0,928 & 0,929 & \textbf{0,982} \\
UNSW & 0,798 & 0,836 & \textbf{0,873}  \\
\hline
\end{tabular}
    
\end{table}

\section{Conclusion}
\label{sec:conclusions}
In this paper we present a new method for data stream anomaly detection called Incremental Quantum Measurement Anomaly Detection (InQMAD). The new method uses the adaptive Fourier feature to map the data to a higher space and density matrices to capture the density estimate. The new method has an initial linear training complexity in terms of the number of training data points. In addition, it has a linear update complexity, which makes it suitable for streaming problems. A systematic evaluation has been performed against 12 state-of-the-art methods on 12 streaming datasets, showing similar performance. We give a theoretical guarantee that the update stage of the algorithm generates a valid density matrix. In addition, an ablation study shows the importance of the new empirical random sampling applied to the adaptive Fourier features and the importance of the update memory parameter $\alpha$. For future research, we will use a dimensionality reduction step such as principal component analysis or autoencoders to help with the curse of dimensionality in massive datasets.

\bibliography{anomaly_detection_2022}
\bibliographystyle{splncs04}

\section{Supplemental material}

\begin{prop}

The resulting matrix  $\rho_{t+1} = (1-\alpha) \cdot \rho_{t} + \alpha \cdot \phi({\bm{x}_{t+1}})\phi^T({\bm{x}_{t+1}})$ in Equation \ref{eq:density_matrix_update} is a valid density matrix of the form $\rho_{t+1} = \sum_{i=1}^{t+1} q_i \phi({\bm{x}_{i}})\phi^T({\bm{x}_{i}})$ with $q_1 =  (1-\alpha)^{t+1}$, $q_i = (1-\alpha)^{t-i+1} \cdot \alpha \text{ } \forall i \in \{1, \cdots, t\}$ and $\alpha \in [0,1]$.
\end{prop}
\begin{proof}
   $\rho_{t+1}$ is a valid density matrix if  $(1-\alpha)^{t+1} + \sum_{i=1}^{t} (1-\alpha)^{t-i+1} \cdot \alpha = 1$. We give a proof by induction on t.
    
\textit{Base Case: } for $t=1$, define $\rho_t = \phi({\bm{x}_{1}})\phi^T({\bm{x}_{1}})$.
for $t=2$, define $q_1 = (1-\alpha)$ and $q_2 = \alpha$ then $q_1 + q_2 = 1$ 

\textit{Induction Step: } Show that for every $t \geq 2$, if $\rho_t$ holds, then $\rho_{t+1}$ also holds.

Define 
\begin{align}
    p_{t} &=  (1-\alpha)^{t} + \alpha \cdot \sum_{i=1}^{t-1} (1-\alpha)^{t-i+1}
\end{align}
Using $\sum_{k=1}^n x^k = (x-x^{n+1})/(1-x)$, then 
\begin{align}
    p_{t} =  (1-\alpha)^{t} + \alpha \cdot (\frac{(1-\alpha) - (1-\alpha)^{t }}{1-(1-\alpha)})  = 1
\end{align}
Now, for t+1 : 
\begin{align}
    q_{t+1} &=  q_1 \cdot \left((1-\alpha)^{t} + \alpha \cdot \sum_{i=1}^{t-1} (1-\alpha)^{t-i+1}\right) + q_2\\
            &= (1-\alpha) \cdot \left((1-\alpha)^{t} + \alpha \cdot \sum_{i=1}^{t-1} (1-\alpha)^{t-i+1}\right) + \alpha \\
            &= (1-\alpha)^{t+1} + \alpha \cdot \sum_{i=1}^{t} (1-\alpha)^{t-i+1} + \alpha \\
            &= (1-\alpha)^{t+1} + \alpha \cdot \sum_{i=0}^{t+1} (1-\alpha)^{t-i+1} \\
            &= 1
\end{align}
\end{proof}

\subsection{Dataset Description}

We present a summarized description of the datasets used in the experimental setup to apply the algorithms on.

\begin{itemize}
    \item Cardio \cite{Rayana2016}: consists of measurements of fetal heart rate, where the original classes are normal, suspect, and pathologic; to adapt it to outlier detection, the normal class formed the inliers and the pathologic class is downsampled and labeled as outliers, while the suspect class is discarded.
    \item Ionosphere \cite{Rayana2016}: originally a binary classification dataset from UCI ML repository, the ‘bad’ class is considered as outliers and the ‘good’ class as inliers.
    \item Mammography  \cite{Rayana2016}: an open dataset about breast calcification, for outlier detection tasks the minority class of `calcification' is considered as outliers and the `non-calcification' class as inliers.
    \item Pima \cite{Rayana2016}: also from UCI ML repository, it includes data of female Indian patients with the objective of classifying them as diabetic or healthy.
    \item Satellite \cite{Rayana2016}: derived from the Statlog dataset from UCI ML repository, it is a multiclass dataset. For anomaly detection, the smallest three classes (2, 4, 5) are combined to form the outliers, while all the other classes are combined to form the inliers. 
    \item Satimage \cite{Rayana2016}: coming from the previous Satellite dataset, here class 2 has been downsampled and considered as outliers, while the other classes are labeled as normal data. This dataset and the latter came from ODDS virtual library. 
    \item Synthetic: this dataset was created in \cite{Bhatia2022} to analyze if their method could afford sudden changes in data distribution. For this, the authors of the dataset combined two sinusoidal waves, and contaminated 10\% of the samples by adding Gaussian noise to simulate anomalous data.
    \item KDD99 \cite{data_kdd}: one of the best-known datasets in anomaly detection, the original dataset contains 34 numerical dimensions and 7 categorical dimensions, that were transformed using one-hot encoding to obtain a dataset of 121 dimensions. We treat normal data as outliers in this experiment, given the fact that only 20\% of all records are labeled as normal.
    \item NSL \cite{data_nsl}: coming from the previous KDD dataset, it adds some dimensions and solves redundant and duplicate records.
    \item Cover \cite{Rayana2016}: originally a multiclass dataset from UCI ML repository, it is used to predict forest cover type from wilderness areas in Colorado. Instances from class 2 are considered as normal and instances from class 4 are labeled as anomalies. Instances from other classes are omitted.
    \item DoS \cite{data_dos}: this dataset was created by the Canadian Institute of Cybersecurity. Each record corresponds to a network package, and they were captured from simulations of normal network traffic and synthetic attacks.
    \item UNSW \cite{data_unsw}: created by ACCS (an Australian institute of computer science and cybersecurity), it contains both real network normal activities and synthetic attacks. Originally it included nine types of attacks. It has 13\% of anomalies.
\end{itemize}

\subsection{Parameters}

The winning combinations of parameters for our method are presented in Table \ref{winning}. For each dataset, their respective parameters are presented as an array.

The parameters selected to run the baseline methods were the same for all datasets, and follow the recommended values in the original proposal of each method. Their values are presented in Table \ref{otherparams}.

\begin{table}[ht!]
\centering
\caption{Best combinations of InQMAD parameters for each dataset}
\label{winning}

\begin{tabular}[l]{p{0.08\textwidth}|p{0.19\textwidth}}
\hline
\multirow{2}{3em}{Dataset} & Winning Parameters \\ & [$n$, $lr_{base}$, sigma $\sigma$, alpha $\alpha$] \\
\hline

Cardio & [256  0,001  3,0  0,99] \\
Cover & [1000  0,01  2,0  0,05] \\
DoS & [2048  0,001  0,11  0,40] \\
Ionosphere & [100  0,001  0,9  0,40] \\
KDD & [5000  0,001  2,0  0,5] \\
Mammo & [512  0,001  4,0  0,80] \\
NSL & [2000  0,01  1,25  0,25] \\
Pima & [64  0,001  0,5  0,95] \\
Satellite & [128  0,001  0,7  0,04] \\
Satimage & [256  0,0001  0,8  0,005] \\
Syn & [64  0,01  0,1  0,04] \\
UNSW & [2000  0,001  2,5  0,6] \\
\hline

\end{tabular}
\end{table}

\begin{table}[ht!]
\centering
\caption{Best parameters for baseline methods}
\label{otherparams}

\begin{tabular}[l]{p{0.075\textwidth}|p{0.39\textwidth}}
\hline
Method & Winning Parameters \\
\hline

STORM & $window\_size=10000, max\_radius=0.1$ \\
HS-Tree & $window\_size=100, num\_trees=25, \newline max\_depth=15,  initial\_window\_X=None$ \\
iForestASD & $window\_size=100, n\_estimators=25, \newline
anomaly\_threshold=0.5, drift\_threshold=0.5$ \\
RS-Hash & $sampling\_points=1000, decay=0.015, \newline
num\_components=100, num\_hash\_fns=1$ \\
RCF & $num\_trees=4, shingle\_size=4, \newline tree\_size=256$ \\
LODA & $num\_bins=10, num\_random\_cuts=100$ \\
Kitsune & $max\_size\_ae=10, learning\_rate=0.1, \newline hidden\_ratio=0.75, grace\_feature\_mapping=0.1$ \\
DILOF & $window\_size = 400, K = 8, \newline thresholds = [0.1, 1.0, 1.1, 1.15, 1.2, 1.3, 1.4, 1.6, 2.0, 3.0]$ \\
xStream & $projection\_size=50, number\_chains=50, \newline depth=10, rowstream=0, nwindows=0, \newline scoring\_batch\_size=100000$ \\
Mstream & $alpha = 0.85$ \\
Ex. IF & $n\_trees=200, sample\_size=256, \newline limit=None,
Extension\_Level=1$ \\
MemStream & $memory\_length=[4,16,32,64,128,256,1024,2048], \newline update\_threshold=[10, 1, 0.1, 0.01, 0.001, 0.0001]$ \\
\hline

\end{tabular}
\end{table}

\subsection{Statistical Tests}

Using the Table \ref{results} as starting point, the  Friedman test, a well-known statistical method to compare different populations or groups, was applied in order to determine whether there are statistically significant differences between the different methods. Friedman test uses the following formula:

$$
Q = \left [ \frac{12}{Nk(k+1)} \sum_{j=1}^{k} R_j^2 \right ] - 3N(k+1)
$$

where $N$ is the number of datasets, $k$ is the number of algorithms, and $R_j^2$ is the squared sum of the observations for each particular algorithm. Given a confidence value of $0.05$ and the p-value given by $\textbf{P}[\chi_{n-1}^2 \geq Q]$, if the p-value is lower than the confidence value, there is statistically significant evidence supporting that the methods are different. Applying the test returns a p-value of $3.06\times10^{-14}$, clearly showing that such difference does exist.

Friedman test does not indicate which of the methods are responsible for this difference, so to compare them two by two, we use the Friedman-Nemenyi test, applying it over all pairs of methods. This test generates coefficients for each pair of datasets that indicate if they are different (near to 0) or not (near to 1).

\begin{figure}
    \centering
    \includegraphics[width=2.9in]{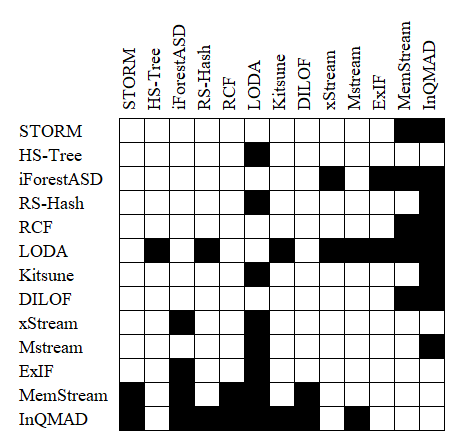}
    \caption{Results of Friedman-Nemenyi test, performed on AUCROC performances of all methods over all datasets.}
    \label{fig:chess}
\end{figure}

Figure \ref{fig:chess} shows the results of Friedman-Nemenyi test, where black squares correspond to pairs of datasets that differ significantly, and white squares correspond to pairs that do not. The most different methods are LODA (due to their poor results) and InQMAD (the best overall method). Other methods that differ notably from others include MemStream (the second best method) and iForestASD (the second worst method). The remaining methods differ with others in only one or two cases.

\end{document}